\documentclass[journal, twoside]{IEEEtran}
\let\labelindent\relax
\usepackage{amsmath}
\usepackage{amssymb}
\usepackage{mathtools}
\usepackage{bm}
\usepackage{tabularx}
\usepackage{ulem}
\usepackage{threeparttable}
\usepackage{booktabs}
\usepackage{diagbox}
\usepackage{array}        
\usepackage{multirow}
\usepackage[table]{xcolor} 
\usepackage{graphicx}
\usepackage{wrapfig} 
\usepackage[lined, ruled, linesnumbered, noend]{algorithm2e}
\usepackage[colorlinks=true, allcolors=blue]{hyperref}

\SetCommentSty{mycommfont}
\usepackage{subcaption}
\usepackage{eqlist}
\usepackage{amsfonts}
\usepackage{enumitem}
\usepackage{lipsum}
\usepackage{stfloats}
\usepackage{booktabs}
\usepackage{multirow}
\usepackage{arydshln}

\usepackage{comment}

\usepackage{multirow}

\usepackage{tikz}
\usetikzlibrary{fit, shapes.misc}

\setlist{nosep}
\setlist[itemize]{left=0pt, topsep=0pt, itemsep=0pt}

\usepackage{CJKutf8}
\definecolor{bleudefrance}{rgb}{0.19, 0.55, 0.91}
\definecolor{awesome}{rgb}{1.0, 0.13, 0.32}
\definecolor{darkgreen}{rgb}{0.0, 0.65, 0.0}
\definecolor{babyblue}{rgb}{0.29, 0.75, 0.93}

\begin{document}
\title{Learning from Planned Data to Improve Robotic Pick-and-Place Planning Efficiency}
\author{Liang Qin$^{1}$, Weiwei Wan$^{1*}$, Jun Takahashi$^{2}$, Ryo Negishi$^{2}$, Masaki Matsushita$^{2}$, and Kensuke Haradan$^{1}$
\thanks{${}^{1}$Graduate School of Engineering Science, The University of Osaka, Japan. ${}^{2}$H.U. Group Research Institute G.K., Japan.}
\thanks{Contact: Weiwei Wan, {\tt\small wan.weiwei.es@osaka-u.ac.jp}}}

\markboth{Under Review by a Robotics Journal, 2025.}
{Qin \MakeLowercase{\textit{et al.}}: Learning from Planned Data to Improve Robotic Pick-and-Place Planning Efficiency}
\maketitle


\bstctlcite{IEEEexample:BSTcontrol}

\begin{abstract} 
This work proposes a learning method to accelerate robotic pick-and-place planning by predicting shared grasps. Shared grasps are defined as grasp poses feasible to both the initial and goal object configurations in a pick-and-place task. Traditional analytical methods for solving shared grasps evaluate grasp candidates separately, leading to substantial computational overhead as the candidate set grows. To overcome the limitation, we introduce an Energy-Based Model (EBM) that predicts shared grasps by combining the energies of feasible grasps at both object poses. This formulation enables early identification of promising candidates and significantly reduces the search space. Experiments show that our method improves grasp selection performance, offers higher data efficiency, and generalizes well to unseen grasps and similarly shaped objects.
\end{abstract}


\section{Introduction} 

\IEEEPARstart{G}{rasp} selection plays a critical role in enabling reliable pick-and-place operations in robotic manipulation. While many existing approaches addressed how to grasp or pick up an object, insufficient consideration of the placement requirements during grasp planning can lead to failures in achieving successful or task-compliant placements. This issue becomes particularly pronounced in industrial settings, where multiple constraints must be satisfied to complete complex manipulation tasks. Typical examples include, but are not limited to, quality inspection tasks where objects must be placed in specific orientations for evaluation; assembly operations requiring precise positioning and orientation for insertion or alignment; and retail scenarios where items must be placed on shelves with logos or labels facing outward. Traditional solutions to these challenges often relied on grasp reasoning frameworks that performed exhaustive search to evaluate grasp candidates based on Inverse Kinematics (IK), grasp stability, and motion feasibility. However, the computational burden increases rapidly with the number of grasp candidates, as each must be evaluated individually across multiple criteria.

To address this issue, we propose a learning-based approach that predicts grasp candidates considering downstream placement feasibility, thereby enabling efficient and reliable planning under real-world constraints.

\begin{figure}[t]
    \centering
    \includegraphics[width=1\linewidth]{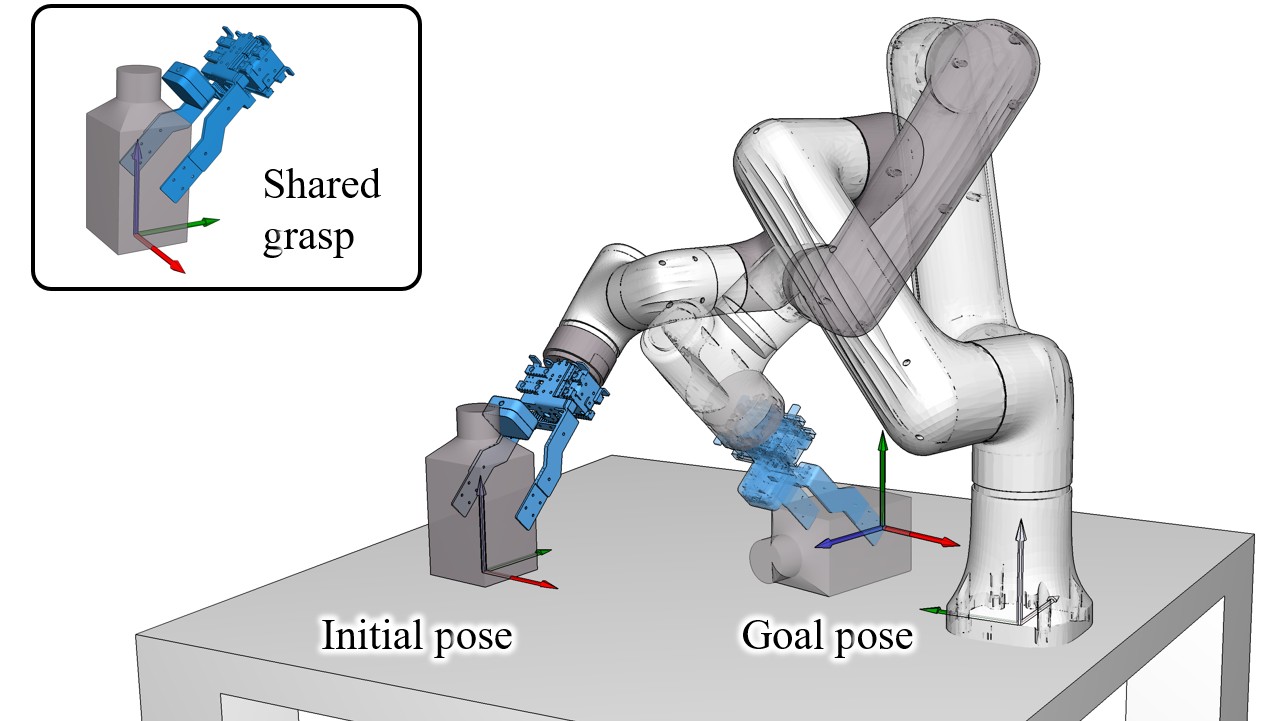}
    \caption{The blue grasp pose on the upper-left corner represents the grasp candidate that can be used to grasp the object at both the initial and goal poses. We refer to such a grasp candidate as a ``shared grasp'' and propose a method to predict it based on the object’s initial and goal poses, thereby accelerating pick-and-place planning.}
    \label{Introduction_picture}
\end{figure}

Our work builds upon the concept of the shared grasp for an object, as introduced in prior studies King et al.~\cite{king2013pregrasp}, Wan and Harada~\cite{wan2015improving}, Xu et al.~\cite{xu2023learning}. A shared grasp refers to a grasp pose defined in the object’s local coordinate frame that remains feasible (i.e., satisfies collision and IK constraints) under both the initial and goal object poses (Fig. \ref{Introduction_picture}). Predicting shared grasps helps narrow down grasp selection, reducing computational overhead in downstream planning.

This work proposes an EBM-based method that efficiently identifies the shared grasps under tabletop constraints. Directly predicting shared grasps in the full state space suffers from high dimensionality, requiring a large amount of training data and incurring significant sampling costs. Instead of direct prediction, we decompose the task into two feasible grasp predictions and combine them. 
Specifically, we introduce an EBM that learns the energy landscape over object poses and grasp poses, assigning low energy to those satisfying IK and collision constraints, and high energy to negative ones. Shared grasps are then identified by composing the energies of two feasible grasps. This structured decomposition not only improves planning efficiency but also reduces training complexity. In addition, our method achieves prediction robustness by leveraging pre-annotated grasp candidates in the object’s local coordinate system, which ensures the predicted grasp poses dependent on the object pose, avoiding compounded errors that typically arise in methods requiring simultaneous estimation of both object and grasp poses. The pre-annotated grasp candidates make the proposed method well-suited for integration into real-world industrial workflows.

We evaluate our method through extensive experiments in both simulation and real-world settings. Results show that our approach significantly improves grasp selection efficiency while maintaining high success rates. The method also demonstrates generalization to unseen grasp poses and new object geometries. A supplementary video showcases the method's integration with visual detection, grasp prediction, and motion planning. In summary, our key contributions are as follows:

\begin{enumerate}[leftmargin=*, nosep]
    \item We propose an EBM-based method for shared grasp prediction under tabletop constraints. The grasp poses to be predicted are selected from a predefined set of grasp candidates in the object's local coordinate frame, which ensures that all outputs remain within a controlled and physically feasible space, avoiding issues commonly seen in unconstrained learning-based methods.
    \item We propose a compositional formulation that decomposes the shared grasp prediction into two feasible grasp evaluations. The formulation significantly reduces the complexity of learning in high-dimensional state spaces and better generalizes capacity compared to other methods.
\end{enumerate}

\section{Related Work} 
\subsection{Effective Pick-and-Place Planning}

Pick-and-place is a fundamental task in robotic manipulation. It requires precise coordination between grasping and placement~\cite{shanthi2024pick, maranci2024enabling}. Such coordination becomes especially important in scenarios like robotic assembly~\cite{9663037} and shelf organization~\cite{costanzo2020manipulation}, where geometric constraints and tight spatial interactions are prevalent.

As the size of the planning space increases, the computation time of traditional search algorithms grows rapidly. The problem becomes more serious in high-dimensional, mixed-integer TAsk and Motion Planning (TAMP), where symbolic and geometric decisions must be jointly resolved. One effective strategy is to reduce the search space based on prior knowledge or learned models. For example, Kim et al.~\cite{kim2019learning} introduced a score-space representation that encodes the performance of constraint subspaces and proposed a policy to select promising regions during planning. Yang et al.~\cite{yang2023sequence} developed PIGINet, a transformer-based model that predicts the feasibility of symbolic task plans before invoking motion planners. Khodeir et al.~\cite{khodeir2023learning} proposed a GNN-based relevance model for best-first stream expansion in PDDLStream, which improved search efficiency in large-scale problems.

A key limitation of symbolic task plans is their lack of geometric awareness, they cannot guarantee that corresponding low-level motions are executable. To address this issue, several methods introduce geometric feasibility predictors based on spatial or visual input. Wells et al.~\cite{wells2019learning} trained an SVM classifier using object position features for box-shaped items. Driess et al.~\cite{driess2020deep} and Xu et al.~\cite{xu2022accelerating} designed networks that take top-down images to estimate whether a trajectory exists for a given pick or place action. Ait Bouhsain et al.~\cite{ait2023learning, ait2023simultaneous} presented AFP-Net and AGFP-Net, which classify the geometric feasibility of discrete pick-and-place actions based on multi-view images, object masks, and grasp mode priors. Park et al.~\cite{park2022scalable} introduced Learned Geometric Feasibility (LGF), a voxel-based model that predicts grasp feasibility from local occupancy and supports logic-geometric planning.

Although these feasibility-aware approaches help filter out invalid candidates and reduce overall planning time, most of them assume a small discrete set of grasp poses. The assumption simplifies the learning and planning process, but limits applicability in precision-critical tasks or when manipulating objects with complex geometries. A sparse candidate set may lead to the exclusion of valid grasps, ultimately reducing the task success rate. Our method addresses this limitation by enabling grasp prediction over a large and diverse candidate set. To this end, our EBM-based method learns a continuous cost landscape over grasp poses defined in the object’s local frame. The resulting energy function provides a differentiable and task-aware scoring mechanism for candidate evaluation, which enables smooth ranking, selection, and prediction.

\subsection{Grasp Selection} 

Unlike grasp generation from scratch, grasp selection aims to identify feasible grasps from a predefined candidate set, given task-specific constraints such as object shape, object pose, and kinematic feasibility. Many previous studies have addressed grasp selection specifically for the picking phase. Herzog et al.~\cite{herzog2012template} and Chen et al.~\cite{chen2022category} retrieved grasps by comparing the geometry of target objects with that of known instances. Other approaches relied on learning-based models to extract grasp-relevant features directly from object geometry~\cite{gualtieri2016high, van2020learning}. More recently, Qian et al.~\cite{qianthinkgrasp} introduced a method that uses a large language model to select grasps from a predefined candidate set in a semantically informed context.

In tasks that involve both picking and placing, grasp selection becomes a core component of the overall planning pipeline. He et al.~\cite{he2023pick2place} selected grasp by maximizing placement affordance, ensuring feasibility under both the initial and goal object poses. Xu et al.~\cite{xu2025grasp} proposed a reinforcement learning framework that encodes both initial and goal states and learns grasp selection policies in an end-to-end manner. For more complex regrasping problems, Wan et al.~\cite{wan2019preparatory} and Cheng et al.~\cite{cheng2022learning} represented grasp-object configurations as nodes in a graph and performed search to identify feasible regrasp sequences. Xu et al.~\cite{xu2022efficient} extended this idea to hierarchical planning by jointly optimizing grasp pairs and intermediate object poses, guided by a learned cost estimator.

While these methods typically model grasp selection as a joint optimization over initial and goal poses, our approach adopts a different perspective. We treat the placing process as a reverse picking process, and formulate the joint selection problem into two independent per-pose evaluations. The formulation is motivated by two observations. First, the majority of candidate grasps are infeasible across arbitrary pose pairs, which creates an extremely imbalanced distribution of positive and negative training data. Second, learning over the full joint space requires a large amount of data and computational cost. By decoupling the problem, our method enables the efficient composition of feasible grasp sets and supports fast, modular shared grasp reasoning.

\section{Modeling Feasible Grasps Using EBM}
\label{EBM_modeling}
The traditional pick-and-place workflow begins with grasp planning by evaluating sample candidates based on stability, feasibility, and kinematic constraints. Then, probabilistic motion planning is employed to identify a collision-free path, and iterative refinement is performed to establish the optimal trajectory. This entire procedure tends to be time-consuming and relatively slow. As the number of grasp candidates increases, the overall search process becomes exponentially more time-consuming. To expedite the entire process, we are inspired by the work of \cite{park2022scalable} about considering the geometric feasibility of different grasp modes in advance to reduce the inference time of motion planning and develop a learning method to predict the shared grasps, which are defined in the object's local coordinate frame and remain feasible under both the initial and goal object poses. The predicted shared grasps can serve as efficient traversal candidates in pick-and-place planning. By transforming each shared grasp from the object’s local frame to the world frame under both the initial and goal poses, we may solve the corresponding IK problems to obtain feasible robot configurations and perform motion planning between the configurations to generate robot trajectories.

Mathematically, given the initial and goal object poses 
\(\boldsymbol{T}_{\mathrm{init}}, \boldsymbol{T}_{\mathrm{goal}} \in SE(3)\), 
and a set of candidate grasp \(\boldsymbol{\mathcal{G}}_0 = \{ (\boldsymbol{g}, w) \}\), 
where \(\boldsymbol{g} \in SE(3)\) represents the gripper's pose in the object's canonical coordinate frame 
\(\boldsymbol{T}_0 = \boldsymbol{I}_{4 \times 4}\), 
and \(w \in \mathbb{R}\) denotes the corresponding normalized gripper width (i.e., finger opening),  we aim to predict the subset 
\(\boldsymbol{\mathcal{G}}_{\mathrm{shared}} \subseteq \boldsymbol{\mathcal{G}}_0\) that remain feasible when transformed under both the initial and goal object poses. In detail, a grasp \((\boldsymbol{g}, w) \in \boldsymbol{\mathcal{G}}_0\) is considered a \emph{shared grasp}
if both the transformed grasp poses $(\boldsymbol{T}_{\mathrm{init}} \cdot \boldsymbol{g}, w)$ and $(\boldsymbol{T}_{\mathrm{goal}} \cdot \boldsymbol{g}, w)$ satisfy the IK and collision constraints. To predict the shared grasps, our method adopts a two-stage learning approach. We first train a model to predict, for a given object pose $\boldsymbol{T}$, whether the transformed grasp pose $(\boldsymbol{T} \cdot \boldsymbol{g}, w)$ satisfies the feasibility constraints. This per-pose feasibility prediction is then extended to shared grasp prediction by evaluating grasp candidates under both $\boldsymbol{T}_{\mathrm{init}}$ and $\boldsymbol{T}_{\mathrm{goal}}$. In the remainder of this section, we focus on modeling the feasible grasp prediction. In the next section, we will discuss combining the predictions under different object poses to construct the shared grasp set.

We use EBM to model the feasible grasp. An EBM defines an energy function over object and grasp pairs and assigns lower energy to more feasible ones. Unlike traditional discriminative models that directly predict a mask vector, an EBM fits the joint probability of object and grasp using
\begin{equation}\label{eq_boltz} p_\phi(x) = \frac{\exp(-E_\phi(x)/t)}{Z_\phi}, \end{equation}
where $x$ denotes the input variables. $E_\phi(x)$ is a learnable energy function parameterized by $\phi$, $t$ is the Boltzmann temperature constant and constant $Z_\phi$ is defined as
\begin{equation} Z_\phi = \int \exp(-E_\phi(x) /t) dx. \end{equation}

It serves as a normalization constant to ensure that $p_\phi(x)$ integrates to one. In practice, $Z_\phi$ is often intractable to compute when the output space is continuous or high-dimensional. However, in our setting, both the input and output spaces are discretized, which allows this integral to be approximated by a summation over a finite set of candidates. In the context of grasp planning, to facilitate training, here we set $x=[\boldsymbol{T}, \boldsymbol{g}, w]$, 
The goal of the EBM is to learn an energy function \(E_{\phi_f}(\boldsymbol{T}, \boldsymbol{g}, w)\) that assigns low energy to feasible grasp -- those that satisfy IK and collision constraints -- under the given object pose. The energy function \(E_{\phi_f}(\boldsymbol{T}, \boldsymbol{g}, w)\) is implemented as a neural network that takes as input the concatenation of a pose encoding of \(\boldsymbol{T}\), a grasp pose \(\boldsymbol{g}\), and gripper width \(w\). It is trained to output a scalar energy value.

To train the energy network, we adopt a Negative Log-Likelihood (NLL) loss function that encourages the model to assign low energy to feasible samples 
\begin{equation}
\mathcal{L}_{\text{nll}} = \mathbb{E}_{\boldsymbol{T}_j} \left[ \mathbb{E}_{(\boldsymbol{g}, w)\,\mathrm{feasible\,for}\,\boldsymbol{T}_j} \left[ \frac{E_{\phi_f}(\boldsymbol{T}_j, \boldsymbol{g}, w)}{t} \right] \right] + \log Z_{\phi_f},
\label{eq_nll}
\end{equation}
where the partition function \(Z_{\phi_f}\) is approximated as
\begin{equation}
Z_{\phi_f} = \sum_{\boldsymbol{T}_j} \sum_{(\boldsymbol{g}, w)} \exp\left(-\frac{E_{\phi_f}(\boldsymbol{T}_j, \boldsymbol{g}, w)}{t}\right).
\label{eq_zphi}
\end{equation}
The \(\mathbb{E}\) in equation~\eqref{eq_nll} represents expectational computations. They are taken over all object poses \(\boldsymbol{T}_j\) in the training set, and the set of grasp candidates \((\boldsymbol{g}, w)\) that are feasible for each \(\boldsymbol{T}_j\). The summations in equation~\eqref{eq_zphi} are taken over all object–grasp pairs encountered during training, regardless of feasibility.

Meanwhile, we use a contrastive energy loss defined as
\begin{equation}
\mathcal{L}_{\mathrm{con}} = \mathcal{L}_{+} - \mathcal{L}_{-},
\label{eq_con}
\end{equation}
where
\begin{equation}
\left\{
\begin{aligned}
\mathcal{L}_{+} &= \mathbb{E}_{\boldsymbol{T}_j} \left[
\mathbb{E}_{(\boldsymbol{g}, w)\,\mathrm{feasible\,for}\,\boldsymbol{T}_j} 
\left[ \frac{E_{\phi_f}(\boldsymbol{T}_j, \boldsymbol{g}, w)}{t} \right]
\right] \\
\mathcal{L}_{-} &= \mathbb{E}_{\boldsymbol{T}_j} \left[
\mathbb{E}_{(\boldsymbol{g}, w)\,\mathrm{infeasible\,for}\,\boldsymbol{T}_j} 
\left[ \frac{E_{\phi_f}(\boldsymbol{T}_j, \boldsymbol{g}, w)}{t} \right]
\right]
\end{aligned}
\right.,
\end{equation}
to assign lower energy to feasible grasps and higher energy to infeasible ones.

In addition, we apply an energy regulation term to prevent divergence
\begin{equation}
\mathcal{L}_{\mathrm{reg}} = \mathcal{L}_{+}^2+\mathcal{L}_{-}^2,
\label{eq_con}
\end{equation}
where
\begin{equation}
\left\{
\begin{aligned}
\mathcal{L}_{+}^2 &= \mathbb{E}_{\boldsymbol{T}_j} \left[
\mathbb{E}_{(\boldsymbol{g}, w)\,\mathrm{feasible\,for}\,\boldsymbol{T}_j} 
\left[ \left(\frac{E_{\phi_f}(\boldsymbol{T}_j, \boldsymbol{g}, w)}{t}\right)^2 \right]
\right] \\
\mathcal{L}_{-}^2 &= \mathbb{E}_{\boldsymbol{T}_j} \left[
\mathbb{E}_{(\boldsymbol{g}, w)\,\mathrm{infeasible\,for}\,\boldsymbol{T}_j} 
\left[ \left(\frac{E_{\phi_f}(\boldsymbol{T}_j, \boldsymbol{g}, w)}{t}\right)^2 \right]
\right]
\end{aligned}
\right..
\end{equation}

The final loss is a weighted sum
\begin{equation}
    \mathcal{L}_{\text{total}} = \mathcal{L}_{\text{nll}} + \mathcal{L}_{\text{con}} + \alpha\mathcal{L}_{\text{reg}},
    \label{EBM_loss_function}
\end{equation}
where $\alpha$ is the proportion constant of the regulation term.

During inference, we follow standard binary classification principles to decide whether the given input is feasible. Specifically, the learned energy model $E_{\phi_f}$ assigns a scalar energy score to each input, and we classify it as feasible if the score is below a threshold $h_f$. 

To determine an optimal $h_f$ value, we perform an optimization procedure using the validation set. Concretely, we evaluate the F1 score on the validation set for a range of threshold values, and select the one that yields the highest F1 score as the optimal value. The F1 score is defined as the harmonic mean of precision and recall:
\begin{equation}
\mathrm{F1}(h) = \frac{2 \cdot \mathrm{Precision}(h) \cdot \mathrm{Recall}(h)}{\mathrm{Precision}(h) + \mathrm{Recall}(h)},
\label{eq_f1}
\end{equation}
where
\begin{align}
\mathrm{Precision}(h) &= \frac{\mathrm{TP}(h)}{\mathrm{TP}(h) + \mathrm{FP}(h)}, \\
\mathrm{Recall}(h) &= \frac{\mathrm{TP}(h)}{\mathrm{TP}(h) + \mathrm{FN}(h)},
\label{eq_pr}
\end{align}
and $\mathrm{TP}, \mathrm{FP}, \mathrm{FN}$ denote the numbers of true positives, false positives, and false negatives, respectively. The optimal $h_f$ value is selected by maximizing the F1 score over the validation set.

\begin{figure}[t]
    \centering
    \includegraphics[width=\linewidth]{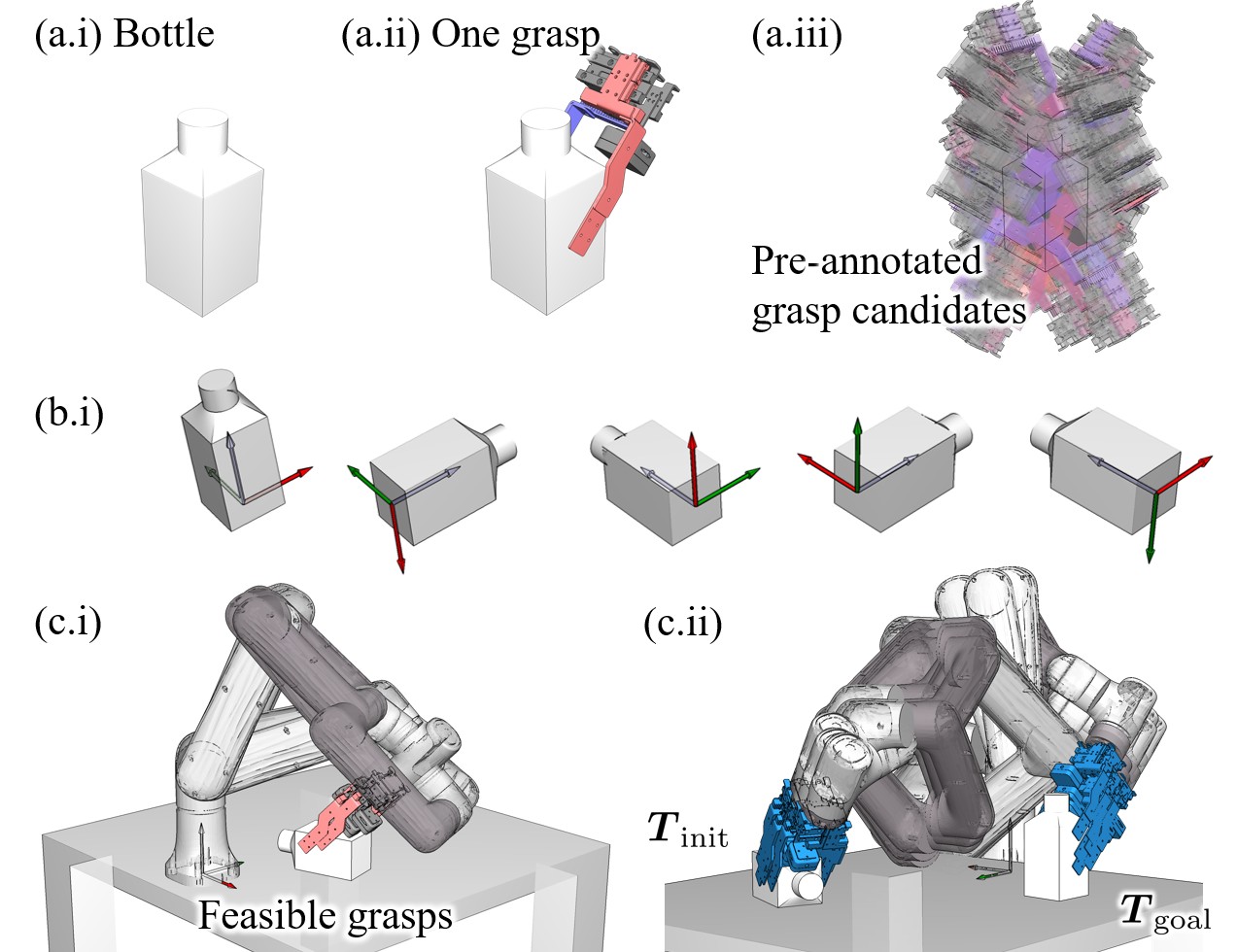}
    \caption{Grasp dataset collection process. (a.i) Target bottle object. (a.ii) The collision-free grasp candidate was generated by sampling the parallel meshes part of the object. (a.iii) The grasp candidate set $\boldsymbol{\mathcal{G}}_0$. (b.i) Stable placements of the object before considering yaw rotations. (c.i) Feasible grasps. (c.ii) Shared grasps.}
    \label{Grasp dataset collection}
\end{figure}

For training data collection, we follow the method proposed by Wan et al.~\cite{wan2015improving} to generate grasp candidates $\{(\boldsymbol{g}, w)\}$ and object poses $\{\boldsymbol{T}_j\}$. The overall pipeline is illustrated in Fig.~\ref{Grasp dataset collection}(a.i–a.iii). Given a 3D object mesh, we uniformly sample surface points and extract antipodal contact pairs as initial grasp candidates. For each pair, we align the gripper to the contact points and generate an initial grasp pose by transforming it to the contact frame with a random rotation around the antipodal axis (Fig.~\ref{Grasp dataset collection}(a.ii)). We then further rotate the gripper around the contact normal and perform collision checking. Grasps that are collision-free and satisfy the force-closure condition are retained (Fig.~\ref{Grasp dataset collection}(a.iii)). This procedure yields $N$ candidate grasps per object.

To generate object poses, we assume a planar tabletop constraint and first compute all statically stable placements of the object (Fig.~\ref{Grasp dataset collection}(b.i)). These base poses are further diversified by randomly sampling planar positions and in-plane yaw angles, which helps effectively cover the SE(2) space. In total, we produce $M$ object poses.

These object-grasp pairs are used to train the EBM with the loss function described previously. Fig.~\ref{Grasp dataset collection}(c.i) shows an example object pose and its corresponding feasible grasps. Fig.~\ref{Grasp dataset collection}(c.ii) illustrates the resulting shared grasp set under a given initial and goal pose pair.

\section{Shared Grasp Prediction} 
\label{sec_comp}

This section introduces the proposed method for predicting the shared grasp set. The prediction is formulated as a joint probability estimation problem using compositional EBMs, and the goal is to identify grasps that are simultaneously feasible under both the initial and goal object configurations. 

In particular, we define the shared grasp probability as a joint distribution over the initial pose $\boldsymbol{T}_{\text{init}}$, goal pose $\boldsymbol{T}_{\text{goal}}$, and a grasp candidate $(\boldsymbol{g}, w)$. Under the compositional EBM framework, this joint distribution is approximated as a product of independent terms:
\begin{equation}
    p(x_1, x_2, ..., x_n) = \prod\nolimits_i p_{\phi}(x_i) \propto \exp\left(-\sum\nolimits_i E_{\phi}(x_i)\right),
    \label{EBM compositional method}
\end{equation}
which, in our context, gives:
\begin{equation}
    p(\boldsymbol{T}_{\text{init}}, \boldsymbol{T}_{\text{goal}}, \boldsymbol{g}, w) 
    = p(\boldsymbol{T}_{\text{init}}, \boldsymbol{g}, w) \cdot p(\boldsymbol{T}_{\text{goal}}, \boldsymbol{g}, w).
\end{equation}

Following the Boltzmann distribution equation \eqref{eq_boltz}, this expression leads to an additive energy representation:
\begin{align}
   p(\boldsymbol{T}_{\text{init}}, &\boldsymbol{T}_{\text{goal}}, \boldsymbol{g}, w) 
    \propto \notag \\ 
    &\exp\left(
        - \left[ 
            E_{\phi_f}(\boldsymbol{T}_{\text{init}}, \boldsymbol{g}, w) +
            E_{\phi_f}(\boldsymbol{T}_{\text{goal}}, \boldsymbol{g}, w)
        \right]
    \right).
\end{align}
The addition implies that given a candidate $(\boldsymbol{g}, w)$, we can compute its total energy as the sum of its energies under both object poses. Accordingly, the prediction of the shared grasp set
is formulated as selecting grasp candidates whose joint energy is lower than a predefined threshold $h_s$:
\begin{equation} 
\boldsymbol{\mathcal{G}}_{\mathrm{shared}} = \left\{(\boldsymbol{g}, w) \mid E_{\phi_f}(\boldsymbol{T}_{\text{init}}, \boldsymbol{g}, w) + E_{\phi_f}(\boldsymbol{T}_{\text{goal}}, \boldsymbol{g}, w) < h_s\right\}.
\end{equation}

It is important to note that $h_s$ differs from the feasible grasp threshold $h_f$ used in the previous section. In the joint framework, we evaluate the summed energy over both poses and apply a single threshold $h_s$ to determine shared grasps without per-pose feasibility prediction. The selection of $h_s$ also follows the F1-based criterion defined in equations~\eqref{eq_f1}$\sim$\eqref{eq_pr}, i.e., maximizing the F1 score over a validation set.

However, since shared grasp prediction involves evaluating grasp feasibility under both initial and goal poses, the computation of TP, FP, and FN must be based on grasp sets that are truly shared across poses. The dataset used to train the EBM model only involved feasible grasp datasets. It lacked information about shared grasps. For this reason, we additionally synthesize a ground-truth shared grasp dataset by randomly sampling object pose pairs $(\boldsymbol{T}_{\mathrm{init}}, \boldsymbol{T}_{\mathrm{goal}})$, analytically evaluating their corresponding feasible grasp sets based on $\boldsymbol{G}_0$, and then computing the intersection of the feasible sets. The resulting dataset, denoted as $\{\boldsymbol{T}_{\mathrm{init}}, \boldsymbol{T}_{\mathrm{goal}}, \boldsymbol{g}, w \}$, provides supervision for evaluating classification performance under joint energy and is used to determine the threshold $h_s$.

\begin{figure}[htbp]
    \centering 
    \includegraphics[width=1\linewidth]{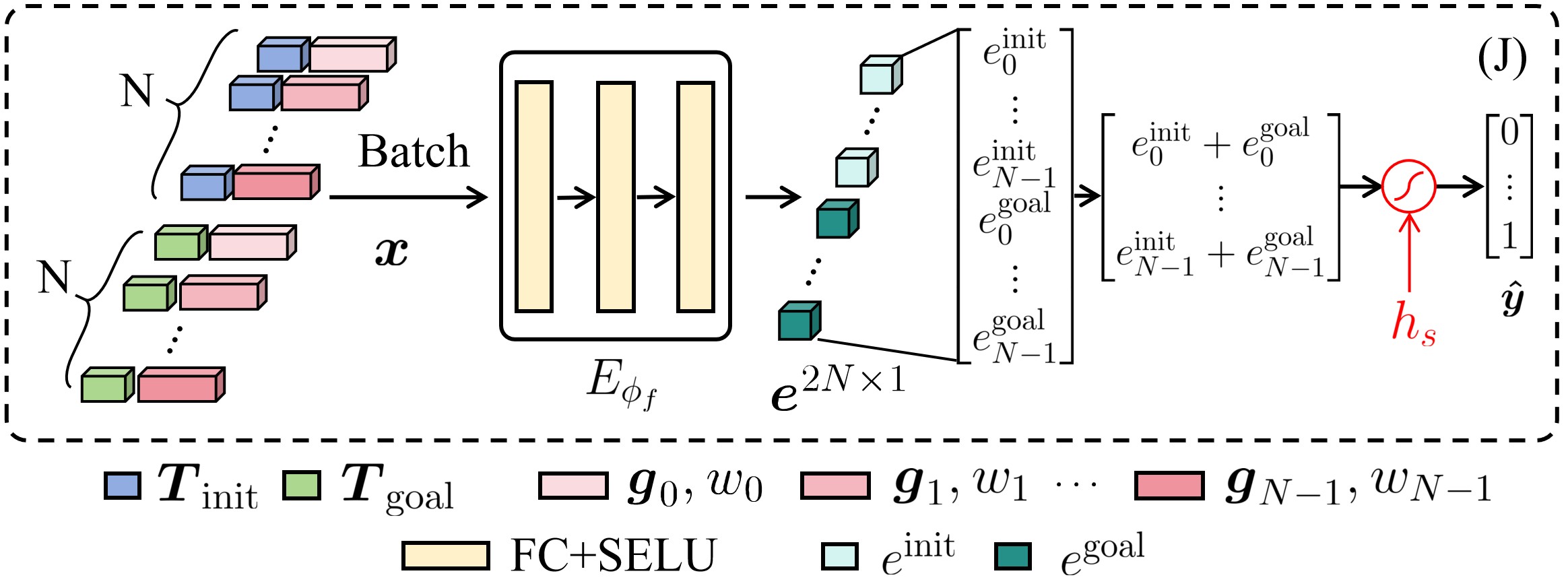}
    \caption{Proposed method uses an EBM~$E_{\phi_f}$ to independently model the energy values of pre-annotated grasp candidates under each object pose and sum up corresponding energy items to jointly estimate the energy value of the shared grasp. Then, the method employs a binary classification strategy to determine whether each grasp qualifies as a shared one.}
    \label{Prediction_method_J}
\end{figure}

To ensure efficient inference, we utilize batch processing. All input tuples $[\boldsymbol{T}_{\text{init}}, \boldsymbol{g}, w]$ and $[\boldsymbol{T}_{\text{goal}}, \boldsymbol{g}, w]$ for $N$ grasp candidates are packed into a single tensor and passed through the energy model in one forward pass. The resulting $2N$ energy values are classified using the threshold $h_s$, yielding a binary mask $\hat{y}$ used to extract the final shared grasp set: $\boldsymbol{\mathcal{G}}_{\text{shared}} = \boldsymbol{\mathcal{G}}_0[\hat{y} = 1]$. The overall procedure is illustrated in Fig.~\ref{Prediction_method_J}.

It is worth noting that the above method represents one possible implementation of shared grasp prediction. In addition to this joint estimation approach, we also consider two variations: a direct prediction method that models shared grasp energy in a single step, and a logical conjunction method that classifies shared grasps based on per-pose feasibility. These variations are described in detail in Appendix and are quantitatively compared with the proposed approach in Section~\ref{sec_exp}. We highlight the joint estimation method in the main text due to its principled formulation and its superior performance observed in experiments.

\section{Experiments and Analysis} 
\label{sec_exp}
We use a 6-DOF Dobot Nova2 robotic arm equipped with a two-finger gripper for the experiments. The training data are collected in the simulation environment illustrated in Fig. \ref{Introduction_picture}. The sampling space is constrained to $x \in [-0.45, 0.45]~\mathrm{m}$, $y \in [0.1, 0.6]~\mathrm{m}$, and $\theta \in [0, 2\pi]~\mathrm{rad}$. The sampling resolution is set to $0.001~\mathrm{m}$ for position and $0.01~\mathrm{rad}$ for orientation. The computational setup used for the experiments consisted of an Intel 13th Gen Core i9-13900KF processor with 64 GB of RAM and an NVIDIA RTX 4090 GPU. Our EBM is implemented using a three-layer fully connected neural network with SELU activation functions. Training is performed with a batch size of 1024, a learning rate of $1 \times 10^{-3}$, a temperature constant of 0.5, and a regularization coefficient $\alpha = 0.2$.

The experiments are divided into two parts. First, we evaluate the overall performance of the proposed joint estimation method (J method). We begin by comparing it with analytical baselines to validate the effectiveness of the EBM-based prediction pipeline. We then compare the J method with two alternative formulations, the direct prediction method (D method) and the logical conjunction method (L method), to examine how different strategies for leveraging EBM outputs affect prediction accuracy and data efficiency under varying training data ratios. Second, we analyze the generalization ability of the J, D, and L methods by testing their performance on unseen grasps and unseen objects.

\subsection{Performance}

\subsubsection{Comparison with Analytical Methods}

We first compare the J method with analytical baselines to evaluate the effectiveness of the EBM-based shared grasp prediction. The experiments are conducted using the bottle-shaped object shown in Fig.~\ref{Grasp dataset collection}. To investigate how the size of the grasp candidate set $\boldsymbol{\mathcal{G}}_0$ affects prediction performance, we pre-generate three sets containing 57, 109, and 352 grasp candidates, respectively. For training the J method, we randomize the object's position and orientation on the table and transform grasp candidates to these poses to generate feasible grasps. For each set of candidate grasps, we collect 75k feasible grasps, among which 50k are used for training, 15k for testing, and 10k for validation (i.e., for synthesizing the ground-truth shared grasp dataset and determining $h_s$). To evaluate the grasp selection strategy, we compare two variants of the J method: (i) $\text{J}^\text{R}$, which randomly selects a grasp from the predicted positive set, and (ii) $\text{J}^\text{O}$, which selects the grasp with the lowest predicted energy among the positive set. As for the analytical baselines, we consider two implementations: (i) R method, which directly samples a grasp uniformly at random from the candidate set without any IK or collision filtering, and (ii) A method, which filters candidates by computing the IK solutions and checking for collisions at both the initial and goal object poses, and then selects a grasp randomly from the intersection of feasible candidates.

Table \ref{tab_analysis_vs_j} presents the comparison results across different methods. The columns are grouped into four major sections, each corresponding to a different size of $\boldsymbol{\mathcal{G}}_0$. The rows are divided into two main parts. The first part reports the results without motion planning, focusing solely on the ability to identify shared grasps. The $\overline{S_\mathrm{G}}$ row reports the success rate of finding a collision-free and IK-feasible shared grasp over 1000 trials for each method. The $\overline{T}$ row shows the average time required to find a successful solution. We can see that the learning-based methods are significantly faster than the analytical baselines. Despite some failures, they achieve consistently high success rates.

The second part of the table includes the results after incorporating motion planning. The $\overline{S_\mathrm{M}}$ row reports the average success rate of motion planning using the predicted shared grasps over 1000 trials. The $\overline{N}$ row shows the average number of candidate shared grasps attempted before finding a feasible motion plan\footnote{If no valid solution is found within 50 attempts, the trial is marked as a failure and excluded from further attempts.}. The results show that the $\text{J}^\text{O}$ method achieves the highest success rate, which may be attributed to the fact that its energy representation captures not only the feasibility of an individual grasp pose, but also the feasibility of its neighboring configurations. This property facilitates the generation of feasible motion plans starting from the predicted grasp.

\begin{table}[!htbp]
\setlength{\tabcolsep}{3pt}
\centering
\caption{Comparison with analytical methods.}
\label{tab_analysis_vs_j}

\begin{threeparttable}
\setlength\tabcolsep{2.7pt}
\begin{tabular}{c|cccc|cccc|cccc}
\toprule
& \multicolumn{4}{c|}{57 grasp candidates} 
& \multicolumn{4}{c|}{109 grasp candidates} 
& \multicolumn{4}{c}{352 grasp candidates} \\
\cmidrule(lr){2-5} \cmidrule(lr){6-9} \cmidrule(lr){10-13}
& R & A & $\text{J}^\text{R}$ & $\text{J}^\text{O}$
& R & A & $\text{J}^\text{R}$ & $\text{J}^\text{O}$
& R & A & $\text{J}^\text{R}$ & $\text{J}^\text{O}$ \\
\midrule
$\overline{S_G}$
& 9.5 & \cellcolor{lime!20}100 & 86.6 & 92.7
& 7.0 & \cellcolor{lime!20}100 & 87.4 & 88.4
& 6.2 & \cellcolor{lime!20}100 & 81.1 & 89.9 \\ \noalign{\vskip 1.5pt}
\hdashline
\noalign{\vskip 2.5pt}
$\overline{T}$
& -- & 5.7 & \multicolumn{2}{c|}{\cellcolor{lime!20}1.7}
& -- & 11.0 & \multicolumn{2}{c|}{\cellcolor{lime!20}3.1}
& -- & 34.1 & \multicolumn{2}{c}{\cellcolor{lime!20}9.2} \\
\midrule
$\overline{S_M}$
& 6.1 & 68.8 & 63.9 & \cellcolor{lime!20}69.9
& 5.3 & 61.8 & 63.2 & \cellcolor{lime!20}67.1
& 4.2 & 58.1 & 59.3 & \cellcolor{lime!20}66.0 \\\noalign{\vskip 1.5pt}
\hdashline
\noalign{\vskip 2.5pt}
$\overline{N}$
& 23.6 & \cellcolor{lime!20}8.5 & 12.0 & 18.2
& 27.0 & \cellcolor{lime!20}14.9 & 15.5 & 19.1
& 26.5 & \cellcolor{lime!20}12.5 & 12.6 & 20.1 \\
\bottomrule
\end{tabular}
\vspace{2pt}
\begin{tablenotes}
  \item[Note 1] R -- Random selection from grasp candidates; 
  \item[Note 2] A -- Random selection from the intersection of grasps that are IK-feasible and collision-free at both initial and goal poses;
  \item[Note 3] $\text{J}^\text{R}$ -- Random selection from the result of J; 
  \item[Note 4] $\text{J}^\text{O}$ -- Selection of the minimum-energy grasp from the result of J;
  \item[Note 5] The best value in each comparison block is highlighted in lime.
\end{tablenotes}
\end{threeparttable}

\end{table}

\subsubsection{Comparison with Other Prediction Variations}

We next compare the proposed J method with two alternative prediction formulations to investigate how different strategies for leveraging the EBM outputs influence grasp prediction accuracy and data efficiency. The first alternative, the D method, directly predicts whether each grasp is shared using a binary classification model trained on ground-truth labels. The second, the L method, applies logical conjunction over two separately trained EBM-based classifiers for initial and goal poses, respectively. Detailed implementation of both methods is provided in the Appendix.

Similar to the previous experiments, we use the bottle-shaped object as the target for evaluation. To ensure a fair comparison across different methods, we fix the size of the grasp candidate set $\boldsymbol{\mathcal{G}}_0$ to 57 pre-generated grasp candidates. For the J and L methods, we randomly sampled object poses and collected 280k feasible grasps. For method D, we randomly sampled object pose pairs and collected 280k shared grasps. Each dataset was split into 200k for training, 50k for testing, and 30k for validation. To better understand the data efficiency and precision of each method, we further subdivide the 200k training data into varying proportions and evaluate the performance under different training data ratios. 

Table~\ref{tab_direct} shows the results. We can see from the table that the J method, even when trained on only 50\% of the 200k training data, outperforms both the D and L methods trained with the full 100\% dataset. This demonstrates the high data efficiency of the J method. We also observe that the D method exhibits comparable performance to the L method. However, the D method inherently requires twice the data collection time, as it needs to generate shared grasp labels over pose pairs, making it the least data-efficient among the three. The L method essentially performs two independent thresholding operations, one applied to the classifier at the initial pose and the other to the classifier at the goal pose. The separation helps filter out more near-threshold false negatives, which likely contributes to the method's higher recall compared to the J method. However, it increases strictness and leads to a reduction in precision. The ``F'' row of the table shows the performance of a network trained solely to predict feasible grasps for reference. From the results, we conclude that the J method offers a relatively balanced performance across recall and precision, which is why we mainly presented it in the main text. Nevertheless, we do not consider the D and L methods to be inferior. In the following generalization experiments, we include all three methods for a comprehensive comparison.

\begin{table}[!htbp]
\setlength{\tabcolsep}{3pt}
\centering
\caption{Comparison of other prediction methods.}
\label{tab_direct}
\begin{threeparttable}
\setlength\tabcolsep{3pt}
\begin{tabular}{l|ccc|ccc|ccc|ccc}
\toprule
& \multicolumn{3}{c|}{100$\%$} 
& \multicolumn{3}{c|}{50$\%$} 
& \multicolumn{3}{c|}{15$\%$} 
& \multicolumn{3}{c}{5$\%$} \\
\cmidrule{2-13}
& P & R & F1 
& P & R & F1 
& P & R & F1 
& P & R & F1 \\
\midrule
J 
& 94.0 & 95.3 & 94.6 
& \cellcolor{lime!20}{94.2} & 91.3 & 93.0 
& 90.0 & 81.9 & 85.7 
& 85.9 & 77.8 & 81.6 \\
D
& \cellcolor{red!20}{92.9} & 95.2 & 94.1 
& 91.2 & 94.4 & 92.8 
& 78.9 & 81.5 & 80.2 
& 70.1 & 80.4 & 74.9 \\
L
& \cellcolor{red!20}{90.7} & 97.3 & 93.8 
& 89.7 & 96.5 & 93.0
& 78.5 & 93.2 & 85.2
& 77.4 & 88.5 & 82.6 \\
\midrule
F
&98.2  &98.7  &98.4  
&98.0  &98.4  &98.2  
&94.5  &96.6  &95.5   
&92.1  &94.0  &93.0   \\
\bottomrule
\end{tabular}

\vspace{2pt} 
\begin{tablenotes}
  \item[Note 1] F -- The EBM model trained only for feasible grasp prediction, without incorporating shared grasp considerations;
  \item[Note 2] The J method trained on 50\% of the training data outperforms both the D and L methods trained with the full 100\% dataset (lime vs. pink).
\end{tablenotes}
\end{threeparttable}

\end{table}

\subsection{Generalization}
\subsubsection{Unseen Grasps}

We were interested in examining whether the EBM-based methods can generalize across the grasp space of the bottle and identify unseen feasible grasp poses. To this end, we trained models using grasp candidate sets of size 57, 83, 109, and 352, and evaluated them on a separate set comprising 922 candidates\footnote{For training, the J and L methods each used 75k feasible grasp samples per candidate set. The D method was trained on 75k shared grasp samples. All datasets were split into 50k for training, 15k for testing, and 10k for validation. The evaluation on the 922-grasp set was conducted using 5k samples.}. The evaluation results are summarized in Table~\ref{tab_unseen}. 

\begin{table}[!htbp]
\setlength{\tabcolsep}{3pt}
\centering  
\caption{Generality to unseen grasp poses.}
\label{tab_unseen}
\begin{threeparttable}
\begin{tabular}{l|ccc|ccc|ccc|ccc}
\toprule
& \multicolumn{3}{c|}{Model - 1} 
& \multicolumn{3}{c|}{Model - 2} 
& \multicolumn{3}{c|}{Model - 3} 
& \multicolumn{3}{c}{Model - 4} \\
\cmidrule{2-13}
& P & R & F1 
& P & R & F1 
& P & R & F1 
& P & R & F1 \\
\midrule
J 
&72.7 &62.7 &67.3
&76.8 &45.5 &57.2
&81.1 &71.2 &75.8
&\cellcolor{lime!20}{89.2} &\cellcolor{lime!20}{86.3} &\cellcolor{lime!20}{87.8} \\
D
& 65.1 & 45.5 & 53.6
& 67.9 & 39.5 & 50.0
& 72.8 & 54.6 & 62.4
& 76.2 & 66.1 & 70.7 \\

L 
& 65.5 & 62.6 & 64.0
& 74.8 & 45.1 & 56.3
& 76.3 & 67.7 & 71.7
& 81.0 & 74.2 & 77.4 \\
\midrule
F
& 97.1 &98.3 &97.7
& 97.9 &98.2 &98.1
& 97.7 &98.3 &98.1
& 97.8 &98.3 &98.2 \\
\bottomrule
\end{tabular}

\begin{tablenotes}
  \item[Note 1] Model 1 $\sim$ 4 trained on 57, 83, 109, 352 grasp candidates;
  \item[Note 2] The best values for shared grasp prediction are highlighted in lime.
\end{tablenotes}

\end{threeparttable}
\end{table}

As shown in Table~\ref{tab_unseen}, increasing the number of training grasp candidates consistently improves generalization to unseen grasp poses. Under the same training conditions, the J method achieves the best performance. In contrast, the D method requires a larger amount of data to perform well, while the L method tends to be overly conservative due to its strict decision criterion.

\subsubsection{Unseen Objects}
\begin{figure}[!htbp]
    \centering
    \includegraphics[width=1\linewidth]{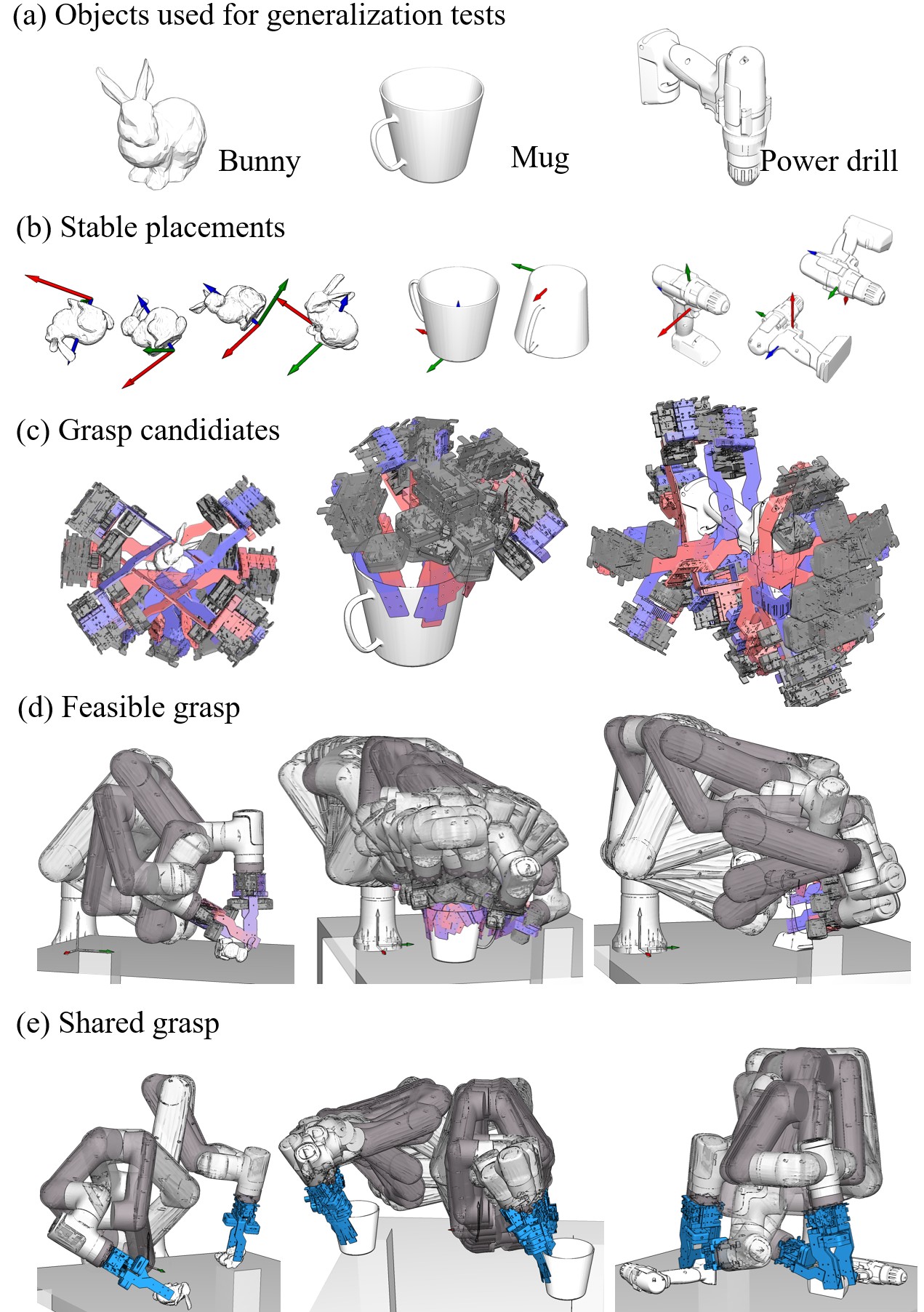}
    \caption{Feasible and shared grasp dataset collection. (a.i) Data collection of the bunny model. (a.ii) Data collection of the mug model. (a.iii) Data collection of the power drill model. (180 grasp candidates for each model)}
    \label{fig_ood_shape}
\end{figure}

We also evaluated the generalization capability of the three methods under variations in object geometry. Specifically, we selected four objects, including a bottle, a bunny, a mug, and a power drill, and prepared 180 grasp candidates for each. Based on these candidates, we collected 150k data per object, including feasibility annotations and shared grasp labels. Fig.~\ref{fig_ood_shape}(a.i-a.iii) illustrates the objects and exemplifies several placements, grasp candidates, feasible grasps, and shared grasps.

To assess cross-object generalization, we trained the models using different combinations of object datasets and tested their performance on the held-out (unseen) ones. Table~\ref{tab_arbitray_obj} summarizes the results. The ``Dataset'' column lists the object combinations used for training. The ``Bunny,'' ``Mug,'' and ``Power drill'' columns report scores of the trained models evaluated on each object. Gray text denotes seen (in-distribution) objects. Black text denotes unseen ones.

\begin{table}[t]
\setlength{\tabcolsep}{4pt}
\centering
\caption{Generality to unseen objects.}
\label{tab_arbitray_obj}
\begin{threeparttable}
\setlength\tabcolsep{3.5pt}
\begin{tabular}{cl|ccc|ccc|ccc}
\toprule
\multirow{2}{*}{} & \multirow{2}{*}{~~~Dataset} 
& \multicolumn{3}{c|}{Bunny} 
& \multicolumn{3}{c|}{Mug} 
& \multicolumn{3}{c}{Power drill} \\
\cmidrule(lr){3-5} \cmidrule(lr){6-8} \cmidrule(lr){9-11}
& & P & R & F1 & P & R & F1 & P & R & F1 \\
\midrule
\multirow{7}{*}{J} & Bt 
& \phantom{0}1.6 &100 &\phantom{0}3.2
& 53.3 &62.1 &57.3
& 23.6 &91.9 &37.6 \\
& Bt$+$Bn
& {\color{gray}88.2} & {\color{gray}97.9} & {\color{gray}92.8}
& 63.0 & 50.3 &55.9
& 26.9 &16.8 & 20.7\\
& Bt$+$Mg
&\phantom{0}1.7 &80.5 &\phantom{0}3.4
& {\color{gray}80.0} & {\color{gray}97.7} & {\color{gray}86.7}
& 20.8 & 87.7 &33.7 \\
& Bt$+$Pd
& \phantom{0}1.6 &98.8 &\phantom{0}3.1
& 32.9 &98.8 & 49.4
& {\color{gray}75.7} & {\color{gray}95.8} & {\color{gray}84.6} \\
& Bt$+$Bn$+$Mg
& {\color{gray}80.6} & {\color{gray}98.8} & {\color{gray}88.8}
& {\color{gray}76.6} & {\color{gray}96.6} & {\color{gray}85.5}
& 29.6 & 47.4 &36.4\\
& Bt$+$Mg$+$Pd
& \phantom{0}1.8 &89.1 &\phantom{0}3.4
& {\color{gray}74.8} & {\color{gray}97.5} & {\color{gray}84.7}
& {\color{gray}73.9} & {\color{gray}96.2} & {\color{gray}83.6} \\
& Bt$+$Bn$+$Pd
& {\color{gray}85.8} & {\color{gray}97.2} & {\color{gray}91.1}
& 47.1 & 54.5 &50.5
& {\color{gray}74.2} & {\color{gray}95.1} & {\color{gray}83.4} \\
\midrule
\multirow{7}{*}{D} & Bt 
& \phantom{0}4.2 & \phantom{0}0.5 & \phantom{0}0.1
&14.7 &\phantom{0}6.7 &\phantom{0}9.2
&21.5 &69.0 &32.8\\
& Bt$+$Bn
& {\color{gray}82.1} & {\color{gray}55.2} & {\color{gray}66.0}
&75.0 & \phantom{0}4.2 &\phantom{0}7.9
&\phantom{0}0.0 &\phantom{0}0.0 &\phantom{0}0.0\\
& Bt$+$Mg
& \phantom{0}0.0 & \phantom{0}0.0 &\phantom{0}0.0
& {\color{gray}78.6} & {\color{gray}85.7} & {\color{gray}82.0}
& \phantom{0}0.0 & \phantom{0}0.0 & \phantom{0}0.0 \\
& Bt$+$Pd
& \phantom{0}6.5 & \phantom{0}2.1 & \phantom{0}3.2
& 47.4 &\phantom{0}0.9 &\phantom{0}1.7
& {\color{gray}74.2} & {\color{gray}75.8} & {\color{gray}75.0} \\
& Bt$+$Bn$+$Mg
& {\color{gray}80.6} & {\color{gray}42.0} & {\color{gray}55.2}
& {\color{gray}76.0} & {\color{gray}82.8} & {\color{gray}79.3}
& \phantom{0}0.0 & \phantom{0}0.0 & \phantom{0}0.0\\
& Bt$+$Mg$+$Pd
& \phantom{0}0.0 &\phantom{0}0.0 &\phantom{0}0.0
& {\color{gray}76.1} & {\color{gray}81.9} & {\color{gray}78.9}
& {\color{gray}71.6} & {\color{gray}70.1} & {\color{gray}70.8} \\
& Bt$+$Bn$+$Pd
& {\color{gray}85.3} & {\color{gray}52.7} & {\color{gray}65.1}
& 36.9 & \phantom{0}0.8 & \phantom{0}1.6
& {\color{gray}71.3} & {\color{gray}73.9} & {\color{gray}72.5} \\

\midrule
\multirow{7}{*}{L} & Bt 
&\phantom{0}6.2 & \phantom{0}7.6 &\phantom{0}6.8
& 84.4 &39.6 & 53.9
&45.5 &17.7 & 25.5 \\
& Bt$+$Bn
& {\color{gray}91.6} & {\color{gray}81.6} & {\color{gray}86.3}
& 88.6 & 17.4 &29.1
& \phantom{0}0.0 &\phantom{0}0.0 &\phantom{0}0.0\\
& Bt$+$Mg
&\phantom{0}5.6 &\phantom{0}0.6 &\phantom{0}1.2
& {\color{gray}80.6} & {\color{gray}92.3} & {\color{gray}86.0}
& \phantom{0}0.0 & \phantom{0}0.0 &\phantom{0}0.0 \\
& Bt$+$Pd
&\phantom{0}2.1 & \phantom{0}0.3 &\phantom{0}0.5
&87.8 &10.1 &18.1
& {\color{gray}78.8} & {\color{gray}88.8} & {\color{gray}83.5} \\
& Bt$+$Bn$+$Mg
& {\color{gray}88.9} & {\color{gray}72.0} & {\color{gray}79.5}
& {\color{gray}79.6} & {\color{gray}91.7} & {\color{gray}85.2}
& \phantom{0}0.0 &\phantom{0}0.0 &\phantom{0}0.0\\
& Bt$+$Mg$+$Pd
& \phantom{0}0.8 & \phantom{0}0.3 & \phantom{0}0.5
& {\color{gray}79.2} & {\color{gray}90.4} & {\color{gray}84.4}
& {\color{gray}77.5} & {\color{gray}85.5} & {\color{gray}81.3} \\
& Bt$+$Bn$+$Pd
& {\color{gray}90.3} & {\color{gray}76.6} & {\color{gray}82.9}
& 72.7 &\phantom{0}6.7 &12.3
& {\color{gray}78.0} & {\color{gray}87.6} & {\color{gray}82.6} \\
\bottomrule
\end{tabular}
\begin{tablenotes}
\item[Note 1] Bt -- Bottle; Bn -- Bunny; Mg -- Mug; Pd -- Power drill;
\item[Note 2] Gray: Results of seen (in-distribution) objects; Black: Results of unseen objects.
\end{tablenotes}
\end{threeparttable}

\end{table}

The results show that all methods generalize reasonably well to the mug object. We attribute this to its geometric similarity to the bottle used in training, as both share comparable feasible grasp distributions. In contrast, generalization to less similar shapes (e.g., bunny and drill) is more challenging.

Among the three methods, the D method performs the worst on unseen objects, suggesting that directly predicting shared grasps lacks robustness under limited training data. The L method shows strong performance on in-distribution data but suffers degraded generalization to the mug when trained jointly with dissimilar object data. This is likely due to interference across shape domains, which affects threshold tuning. The J method, despite slightly lower precision, demonstrates better overall generalization across different objects, making it more suitable for scaling to diverse object categories.

\section{Conclusions and Future Work}
We proposed an EBM-based method to model feasible grasps and predicted shared grasps by composing the learned energies of two feasible picks. Compared to analytical methods, the proposed method achieved a significant speed-up for known grasps while maintaining satisfying accuracy. Moreover, it was able to generalize to unseen grasps and objects. In comparison with other shared grasp prediction methods, the method achieved superior data efficiency and generalization. Experiments were conducted under a simplified tabletop setting to validate and compare the proposed method. Extending it to more complex environments remains an important direction for future work.


\normalem
\bibliographystyle{IEEEtran}
\bibliography{citations.bib}

\begin{thebibliography}{10}
\providecommand{\url}[1]{#1}
\csname url@samestyle\endcsname
\providecommand{\newblock}{\relax}
\providecommand{\bibinfo}[2]{#2}
\providecommand{\BIBentrySTDinterwordspacing}{\spaceskip=0pt\relax}
\providecommand{\BIBentryALTinterwordstretchfactor}{4}
\providecommand{\BIBentryALTinterwordspacing}{\spaceskip=\fontdimen2\font plus
\BIBentryALTinterwordstretchfactor\fontdimen3\font minus \fontdimen4\font\relax}
\providecommand{\BIBforeignlanguage}[2]{{%
\expandafter\ifx\csname l@#1\endcsname\relax
\typeout{** WARNING: IEEEtran.bst: No hyphenation pattern has been}%
\typeout{** loaded for the language `#1'. Using the pattern for}%
\typeout{** the default language instead.}%
\else
\language=\csname l@#1\endcsname
\fi
#2}}
\providecommand{\BIBdecl}{\relax}
\BIBdecl

\bibitem{king2013pregrasp}
J.~E. King, M.~Klingensmith, C.~M. Dellin, M.~R. Dogar, P.~Velagapudi, N.~S. Pollard, and S.~S. Srinivasa, ``Pregrasp manipulation as trajectory optimization.'' in \emph{Robotics: Science and Systems}, 2013.

\bibitem{wan2015improving}
W.~Wan, M.~T. Mason, R.~Fukui, and Y.~Kuniyoshi, ``Improving regrasp algorithms to analyze the utility of work surfaces in a workcell,'' in \emph{IEEE International Conference on Robotics and Automation (ICRA)}, 2015, pp. 4326--4333.

\bibitem{xu2023learning}
P.~Xu, Z.~Chen, J.~Wang, and M.~Q.-H. Meng, ``Learning to predict diverse stable placements for extrinsic manipulation on a support plane,'' \emph{IEEE Transactions on Cognitive and Developmental Systems}, vol.~16, no.~3, pp. 1095--1107, 2023.

\bibitem{shanthi2024pick}
M.~D. Shanthi and T.~Hermans, ``Pick and place planning is better than pick planning then place planning,'' \emph{IEEE Robotics and Automation Letters}, vol.~9, no.~3, pp. 2790--2797, 2024.

\bibitem{maranci2024enabling}
E.~Maranci, S.~D'Avella, P.~Tripicchio, C.~Avizzano \emph{et~al.}, ``Enabling grasp synthesis approaches to task-oriented grasping considering the end-state comfort and confidence effects,'' \emph{IEEE Robotics and Automation Letters}, vol.~9, no.~6, pp. 5695--5702, 2024.

\bibitem{9663037}
H.~Chen, W.~Wan, K.~Koyama, and K.~Harada, ``Planning to build block structures with unstable intermediate states using two manipulators,'' \emph{IEEE Transactions on Automation Science and Engineering}, vol.~19, no.~4, pp. 3777--3793, 2022.

\bibitem{costanzo2020manipulation}
M.~Costanzo, S.~Stelter, C.~Natale, S.~Pirozzi, G.~Bartels, A.~Maldonado, and M.~Beetz, ``Manipulation planning and control for shelf replenishment,'' \emph{IEEE Robotics and Automation Letters}, vol.~5, no.~2, pp. 1595--1601, 2020.

\bibitem{kim2019learning}
B.~Kim, Z.~Wang, L.~P. Kaelbling, and T.~Lozano-P{\'e}rez, ``Learning to guide task and motion planning using score-space representation,'' \emph{The International Journal of Robotics Research}, vol.~38, no.~7, pp. 793--812, 2019.

\bibitem{yang2023sequence}
Z.~Yang, C.~Garrett, T.~Lozano-Perez, L.~Kaelbling, and D.~Fox, ``Sequence-based plan feasibility prediction for efficient task and motion planning,'' in \emph{Robotics science and systems}, 2023.

\bibitem{khodeir2023learning}
M.~Khodeir, B.~Agro, and F.~Shkurti, ``Learning to search in task and motion planning with streams,'' \emph{IEEE Robotics and Automation Letters}, vol.~8, no.~4, pp. 1983--1990, 2023.

\bibitem{wells2019learning}
A.~M. Wells, N.~T. Dantam, A.~Shrivastava, and L.~E. Kavraki, ``Learning feasibility for task and motion planning in tabletop environments,'' \emph{IEEE Robotics and Automation Letters}, vol.~4, no.~2, pp. 1255--1262, 2019.

\bibitem{driess2020deep}
D.~Driess, O.~Oguz, J.-S. Ha, and M.~Toussaint, ``Deep visual heuristics: Learning feasibility of mixed-integer programs for manipulation planning,'' in \emph{IEEE International Conference on Robotics and Automation (ICRA)}, 2020, pp. 9563--9569.

\bibitem{xu2022accelerating}
L.~Xu, T.~Ren, G.~Chalvatzaki, and J.~Peters, ``Accelerating integrated task and motion planning with neural feasibility checking,'' \emph{arXiv preprint arXiv:2203.10568}, 2022.

\bibitem{ait2023learning}
S.~Ait~Bouhsain, R.~Alami, and T.~Simeon, ``Learning to predict action feasibility for task and motion planning in 3d environments,'' in \emph{IEEE International Conference on Robotics and Automation (ICRA)}, 2023, pp. 3736--3742.

\bibitem{ait2023simultaneous}
------, ``Simultaneous action and grasp feasibility prediction for task and motion planning through multi-task learning,'' in \emph{IEEE/RSJ International Conference on Intelligent Robots and Systems (IROS)}, 2023, pp. 2042--2048.

\bibitem{park2022scalable}
S.~Park, H.~C. Kim, J.~Baek, and J.~Park, ``Scalable learned geometric feasibility for cooperative grasp and motion planning,'' \emph{IEEE Robotics and Automation Letters}, vol.~7, no.~4, pp. 11\,545--11\,552, 2022.

\bibitem{herzog2012template}
A.~Herzog, P.~Pastor, M.~Kalakrishnan, L.~Righetti, T.~Asfour, and S.~Schaal, ``Template-based learning of grasp selection,'' in \emph{IEEE International Conference on Robotics and Automation (ICRA)}, 2012, pp. 2379--2384.

\bibitem{chen2022category}
H.~Chen, T.~Kiyokawa, W.~Wan, and K.~Harada, ``Category-association based similarity matching for novel object pick-and-place task,'' \emph{IEEE Robotics and Automation Letters}, vol.~7, no.~2, pp. 2961--2968, 2022.

\bibitem{gualtieri2016high}
M.~Gualtieri, A.~Ten~Pas, K.~Saenko, and R.~Platt, ``High precision grasp pose detection in dense clutter,'' in \emph{IEEE/RSJ International Conference on Intelligent Robots and Systems (IROS)}, 2016, pp. 598--605.

\bibitem{van2020learning}
M.~Van~der Merwe, Q.~Lu, B.~Sundaralingam, M.~Matak, and T.~Hermans, ``Learning continuous 3d reconstructions for geometrically aware grasping,'' in \emph{IEEE International Conference on Robotics and Automation (ICRA)}, 2020, pp. 11\,516--11\,522.

\bibitem{qianthinkgrasp}
Y.~Qian, X.~Zhu, O.~Biza, S.~Jiang, L.~Zhao, H.~Huang, Y.~Qi, and R.~Platt, ``Thinkgrasp: A vision-language system for strategic part grasping in clutter,'' in \emph{2nd CoRL Workshop on Learning Effective Abstractions for Planning}, 2024.

\bibitem{he2023pick2place}
Z.~He, N.~Chavan-Dafle, J.~Huh, S.~Song, and V.~Isler, ``Pick2place: Task-aware 6dof grasp estimation via object-centric perspective affordance,'' in \emph{IEEE International Conference on Robotics and Automation (ICRA)}, 2023, pp. 7996--8002.

\bibitem{xu2025grasp}
K.~Xu, Z.~Zhou, J.~Wu, H.~Lu, R.~Xiong, and Y.~Wang, ``Grasp, see, and place: Efficient unknown object rearrangement with policy structure prior,'' \emph{IEEE Transactions on Robotics}, vol.~41, pp. 464--483, 2025.

\bibitem{wan2019preparatory}
W.~Wan, K.~Harada, and F.~Kanehiro, ``Preparatory manipulation planning using automatically determined single and dual arm,'' \emph{IEEE Transactions on Industrial Informatics}, vol.~16, no.~1, pp. 442--453, 2019.

\bibitem{cheng2022learning}
S.~Cheng, K.~Mo, and L.~Shao, ``Learning to regrasp by learning to place,'' in \emph{Conference on Robot Learning}, 2022, pp. 277--286.

\bibitem{xu2022efficient}
K.~Xu, H.~Yu, R.~Huang, D.~Guo, Y.~Wang, and R.~Xiong, ``Efficient object manipulation to an arbitrary goal pose: Learning-based anytime prioritized planning,'' in \emph{International Conference on Robotics and Automation (ICRA)}, 2022, pp. 7277--7283.

\end{thebibliography}
\section*{Appendix} 

\label{sec_appendix}

\begin{figure}[!htbp]
    \centering 
    \includegraphics[width=1\linewidth]{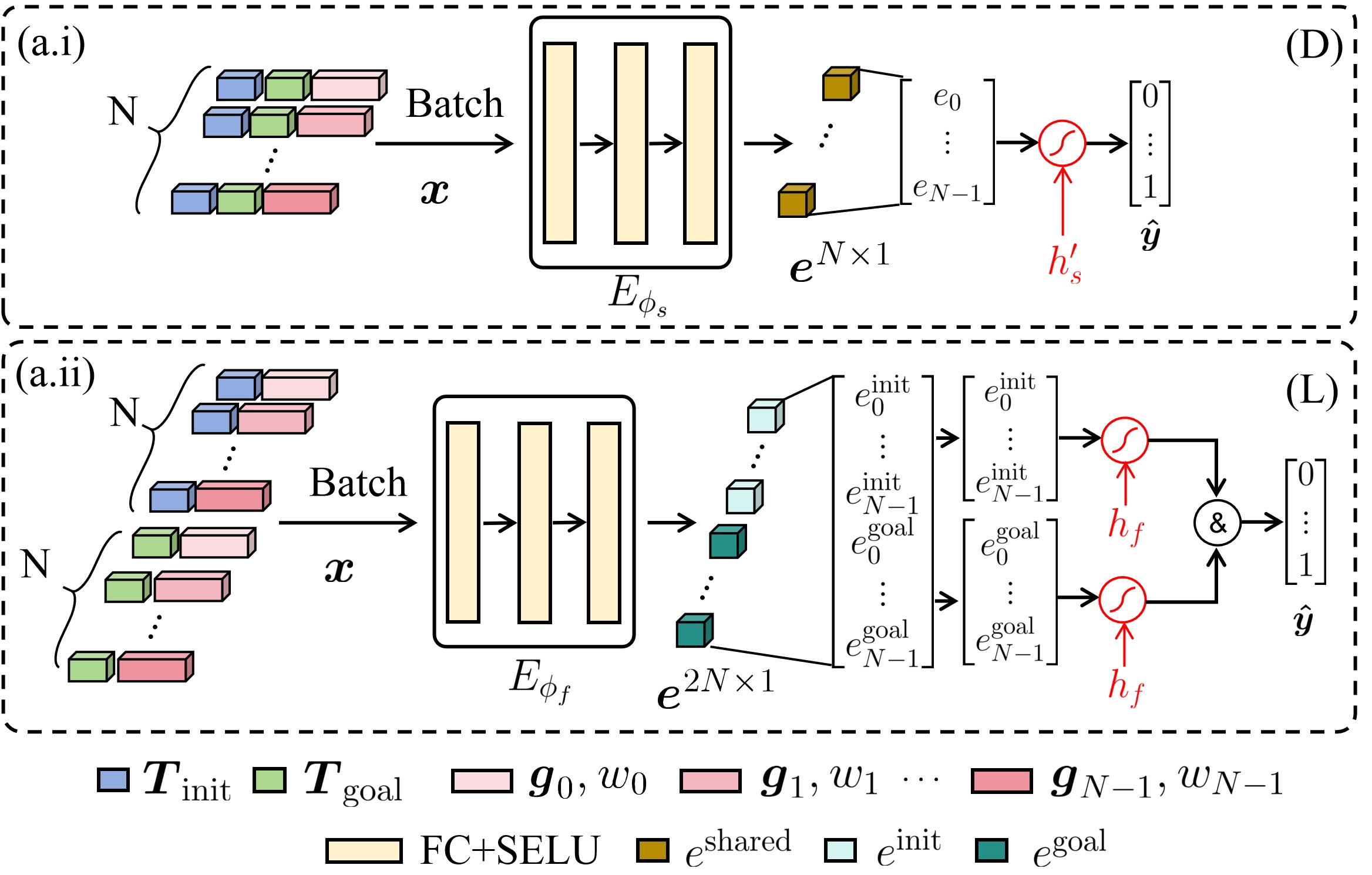}
    \caption{(a.i) Direct Prediction estimates the energy value of shared grasps using an unified EBM model. (a.ii) Logical Conjunction uses an EBM to estimate the feasibility of each object pose and grasp candidate pair, and then employs a Logical AND for predicting the shared ones.}
    \label{Prediction_method}
\end{figure}

We implemented two varied methods for shared grasp prediction: Direct Prediction (D) and Logical Conjunction (L). They served as comparative references in the experiments. Their prediction workflows are illustrated in Fig. \ref{Prediction_method}.

\subsection{Direct Prediction (D Method)}
The D method models the shared grasp distribution directly. It uses an unified EBM $E_{\phi_s}$ to assign low energy to grasp candidates $(\boldsymbol{g}, w)$ that are feasible under both the initial and goal poses. The method takes as input the combined tuple $(\boldsymbol{t}_{\text{init}}, \boldsymbol{t}_{\text{goal}}, \boldsymbol{g}, w)$ and is trained using the same contrastive objective as the feasibility EBM.

During inference, the trained model outputs energy scores for candidate grasps, and a threshold $h_s^\prime$ is applied to classify whether each candidate belongs to the shared grasp set:
\begin{equation}
\boldsymbol{\mathcal{G}}_{\mathrm{shared}} = \left\{(\boldsymbol{g}, w) \mid E_{\phi_s}(\boldsymbol{T}_{\text{init}}, \boldsymbol{T}_{\text{goal}}, \boldsymbol{g}, w) < h_s^\prime \right\}.
\end{equation}
The $h_s^\prime$ is selected by maximizing F1 score on a shared grasp validation set constructed in the same manner as described in Section~\ref{sec_comp}. 

The D method offers a straightforward implementation, but requires collecting supervision from explicitly labeled shared grasps and relies on the model’s ability to learn the implicit joint constraint.

\subsection{Logical Conjunction (L Method)}

The L method classifies shared grasps by explicitly enforcing feasibility thresholding at both the initial and goal poses. A candidate is considered shared if it is feasible under both object poses. The method involves two times of thresholding using $h_f$.



\end{document}